# Text Analysis Tools in Spoken Language Processing

Michael Riley    Richard Sproat

*Room 2d-45*{*4,1*}

*Linguistics Research Department*

*AT&T Bell Laboratories*

*Murray Hill, NJ, 07974, USA*

{riley,rws}@research.att.com

A tutorial presented at the

32nd Annual Meeting of the Association for Computational Linguistics

New Mexico State University, Las Cruces, New Mexico

June 27, 1994

# Outline

- Computational Methods
  - Methods of Inference
  - Implementation
- Incorporation into Language Analysis Modules
  - Language Analysis Modules: Examples
  - Architecture of AT&T Text-to-Speech
    Applications to Speech Recognition
    Applications to Spoken Language Identification

# Computational Methods

- Methods of Inference
  - *Hand-crafted rules*
  - *Statistical Methods: N-Grams*
  - *Statistical Methods: Decision Trees*
  - *Statistical Methods: Decision Lists*
  - *Mixed Methods*
- Methods of Implementation
  - *Weighted Finite-State Acceptors/Transducers*

# Hand-Crafted Rulesets

- Context-free syntactic rewrite rules: $S \Rightarrow NP\,VP$
- Phonological rewrite rules: $C \rightarrow [-voiced]/__\#$
- Tree-to-tree transduction rules:

```
;; input pattern
((E_ADJ (= CAT E_ADJ)
        (MORPH-0 (= LEX "U_S")))
;; output pattern
 (E_ADJ ()
        (MORPH-0
          (& (= LEX "american")
             (= SRC "U_S")))
          (() (& (= cat s_gen)
                 (= sgen (e_adj sgen))))
          (() (& (= cat s_num)
                 (= snum (e_adj snum))))))
```

# Hand-Crafted Rulesets: a Speech Example

Some Rules for foreign name pronunciation in English

**German**

sch $\rightarrow$ ʃ

tʃ $\rightarrow$ č

ei $\rightarrow$ $a^j$

**French**

eau $\rightarrow$ o

Cons $\rightarrow$ Ø/ ___ #

**Japanese**

$$\sigma \rightarrow \acute{\sigma} \; / \; ___ \; \sigma \; \#$$

## Hand-Crafted Rules: Advantages/Limitations

- Advantages:
  - Easy to encode linguistic knowledge directly and precisely
  - Resulting rules are (usually) readily comprehensible to a linguist
- Limitations:
  - Rulesets are often large and complicated:
    - Construction is costly
    - Rule interactions are hard to manage. However, rule development environments — e.g. TWOL (Dalrymple et al., 1987) — are useful here
  - Nonprobabilistic:
    - Systems usually output all possible analyses *without associated weights*
    - Bad for many speech applications

# N-Grams: Basics

- **'Chain Rule' and Joint/Conditional Probabilities:**

  $$P[x_1 x_2 \ldots x_N] = P[x_N | x_1 ... x_{N-1}] P[x_{N-1} | x_1 ... x_{N-2}] \ldots P[x_2 | x_1] P[x_1]$$

  where, e.g.,

  $$P[x_N | x_1 \ldots x_{N-1}] = \frac{P[x_1 \ldots x_N]}{P[x_1 \ldots x_{N-1}]}$$

- **(First–Order) Markov assumption:**

  $$P[x_k | x_1 \ldots x_{k-1}] = P[x_k | x_{k-1}] = \frac{P[x_{k-1} x_k]}{P[x_{k-1}]}$$

- **nth–Order Markov assumption:**

  $$P[x_k | x_1 \ldots x_{k-1}] = P[x_k | x_{k-n} ... x_{k-1}] = \frac{P[x_{k-n} \ldots x_k]}{P[x_{k-n} \ldots x_{k-1}]}$$

# N-Grams: Maximum Likelihood Estimation

Let $N$ be total number of n-grams observed in a corpus and $c(x_1 \ldots x_n)$ be the number of times the n-gram $x_1 \ldots x_n$ occurred. Then

$$P[x_1 \ldots x_n] = \frac{c(x_1 \ldots x_n)}{N}$$

is the maximum likelihood estimate of that n-gram probability.

For conditional probabilities,

$$P[x_n | x_1 \ldots x_{n-1}] = \frac{c(x_1 \ldots x_n)}{c(x_1 \ldots x_{n-1})}.$$

is the maximum likelihood estimate.

With this method, an n-gram that does not occur in the corpus is assigned **zero** probability.

# N-Grams: Good-Turing-Katz Estimation

Let $n_r$ be the number of n-grams that occurred $r$ times. Then

$$P[x_1 \ldots x_n] = \frac{c^*(x_1 \ldots x_n)}{N}$$

is the Good-Turing estimate of that n-gram probability, where $c^*(x) = (c(x) + 1)\frac{n_{c(x)+1}}{n_{c(x)}}$.

For conditional probabilities,

$$P[x_n | x_1 \ldots x_{n-1}] = \frac{c^*(x_1 \ldots x_n)}{c(x_1 \ldots x_{n-1})}, \qquad c(x_1 \ldots x_n) > 0$$

is Katz's extension of the Good-Turing estimate.

With this method, an n-gram that does not occur in the corpus is assigned the backoff probability $P[x_n | x_1 \ldots x_{n-1}] = \alpha P[x_n | x_2 \ldots x_{n-1}]$, where $\alpha$ is a normalizing constant.

# N-Grams: Advantages/Limitiations

- **Advantages:**
  - Captures local, conditional probabilistic information well with adequate data.
  - Simple to use/understand.
  - Efficient implementation.

- **Limitations:**
  - Fails to capture wider-context information.
  - Only limited degree of context generalization (from the back-off), so technique is weak when data is sparse (cf., manual/automatic context clustering).

# Stochastic Part-of-Speech Assignment

- Words may have multiple grammatical parts of speech:

  He/PPS will/MD **table/VB** the/AT motion/NN

  The/AT **table/NN** is/BEZ ready/JJ

  **Can/MD** they/PPSS **can/VB cans/NNS**

- Solution is to use an n-gram model trained on a large corpus of tagged text (Church, 1988; DeRose, 1988), or on an untagged corpus using a dictionary and a reestimation procedure (Kupiec, 1992).

$$argmax \prod_{i=2}^{n-1} \frac{p(word_i|part_i)\, p(part_i|part_{i-1}part_{i-2})}{p(word_i)}$$

# Language-of-Origin Identification for Names

| Name | Language |
|---|---|
| *Vitale* | Italian |
| *Fujisaki* | Japanese |
| *Rodriguez* | Spanish |
| *Blaustein* | German |
| *Andruszkiewicz* | Polish |
| *Perrault* | French |

- Solution: letter trigrams can be used to model the graphotactics of each language (Church, 1986; Vitale, 1991).

# Language-of-Origin Identification for Names

| Trigram probabilities for *Vitale* (Vitale, 1991, p. 265) | | | | |
|---|---|---|---|---|
| Trigram | $p(L_1\|T)$ (Italian) | $p(L_2\|T)$ | $\ldots$ | $p(L_n\|T)$ |
| #vi | .4659 | .0679 | $\ldots$ | .2093 |
| vit | .4145 | .0263 | $\ldots$ | .0000 |
| ita | .7851 | .0490 | $\ldots$ | .0564 |
| tal | .4422 | .1013 | $\ldots$ | .2384 |
| ale | .2602 | .0867 | $\ldots$ | .2892 |
| le# | .3181 | .1884 | $\ldots$ | .0688 |
| *mean* | .4477 | .0866 | $\ldots$ | .1437 |

# Foreign Name Detection in Chinese

| 以 | 博根 | ， | 密德薩斯 | 和 | 孟莫斯 | 三 | 郡 | 為 | 例 |
|---|---|---|---|---|---|---|---|---|---|
| yi3 | bo2−gen1 | | mi4−de2−sa4−si1 | he2 | meng4−mo4−si1 | san1 | jun4 | wei2 | li4 |
| *take* | *Bergen* | | *Middlesex* | *and* | *Monmouth* | *three* | *county* | *for* | *example* |

'... taking the three counties of Bergen, Middlesex and Monmouth, as an example ...'

| 在 | 紐瓦克 | 圖書館 |
|---|---|---|
| zai4 | niu3−wa3−ke4 | tu2−shu1−guan3 |
| *in* | *Newark* | *library* |

'... in Newark library ...'

- Only a couple of hundred characters are at all common in transliterating foreign names, so can build a simple n-gram model modeling these characters (see Sproat et al., 1994).

# Decision Trees: Overview

- **Description/Use:** Simple structure – binary tree of decisions, terminal nodes determine prediction *(cf. "Game of Twenty Questions")*. If dependent variable is categorical (e.g., `red, yellow, green`), called "classification tree", if continuous, called "regression tree".

- **Creation/Estimation:** Creating a binary decision tree for classification or regression involves three steps *(Breiman, et al)*:
  1. *Splitting Rules:* Which split to take at a node?
  2. *Stopping Rules:* When to declare a node terminal?
  3. *Node Assignment:* Which class/value to assign to a terminal node?

# 1. Decision Tree Splitting Rules

Which split to take at a node?

- **Candidate splits considered.**
  - *Binary cuts*: For **continuous** $-\infty \leq x < \infty$, consider splits of form:
  $$x \leq k \quad \text{vs.} \quad x > k, \quad \forall k.$$
  - *Binary partitions*: For **categorical** $x \in \{1, 2, ..., n\} = X$, consider splits of form:
  $$x \in A \quad \text{vs.} \quad x \in X - A, \quad \forall A \subset X.$$

# 1. Decision Tree Splitting Rules – Continued

- **Choosing best candidate split.**
  - *Method 1*: Choose $k$ (continuous) or $A$ (categorical) that minimizes estimated classification (regression) error after split.
  - *Method 2 (for classification)*: Choose $k$ or $A$ that minimizes estimated entropy after that split.

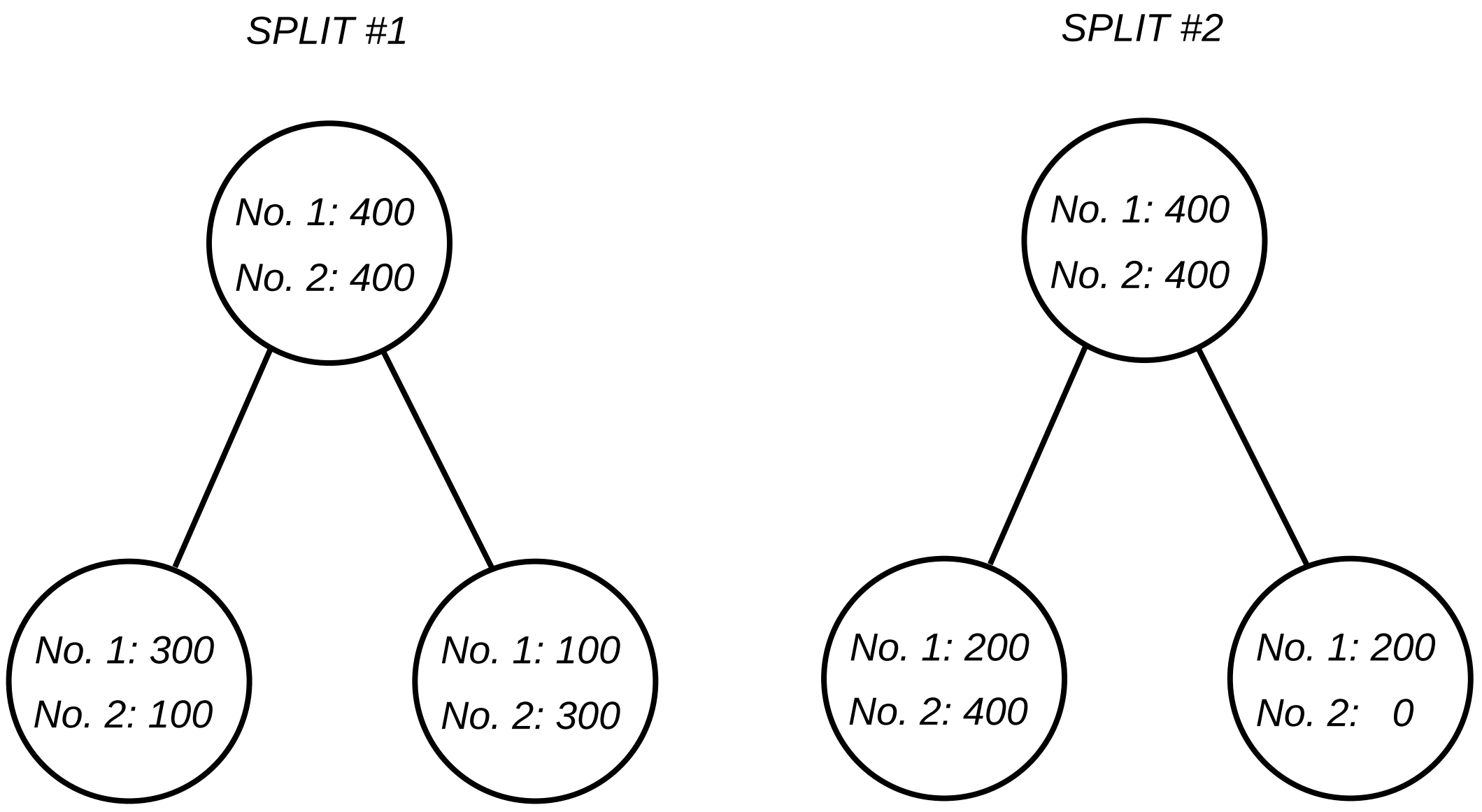

# 2. Decision Tree Stopping Rules

When to declare a node terminal?

- Strategy (*Cost-Complexity pruning*):
  1. Grow over-large tree.
  2. Form sequence of subtrees, $T_0, ..., T_n$ ranging from full tree to just the root node.
  3. Estimate "honest" error rate for each subtree.
  4. Choose tree size with mininum "honest" error rate.
- To form sequence of subtrees, vary $\alpha$ from 0 (for full tree) to $\infty$ (for just root node) in:

$$\min_T \left[ R(T) + \alpha |T| \right] .$$

- To estimate "honest" error rate, test on data different from training data, e.g., grow tree on $9/10$ of available data and test on $1/10$ of data repeating 10 times and averaging (*cross-validation)*.

# End of Declarative Sentence Prediction: Pruning Sequence

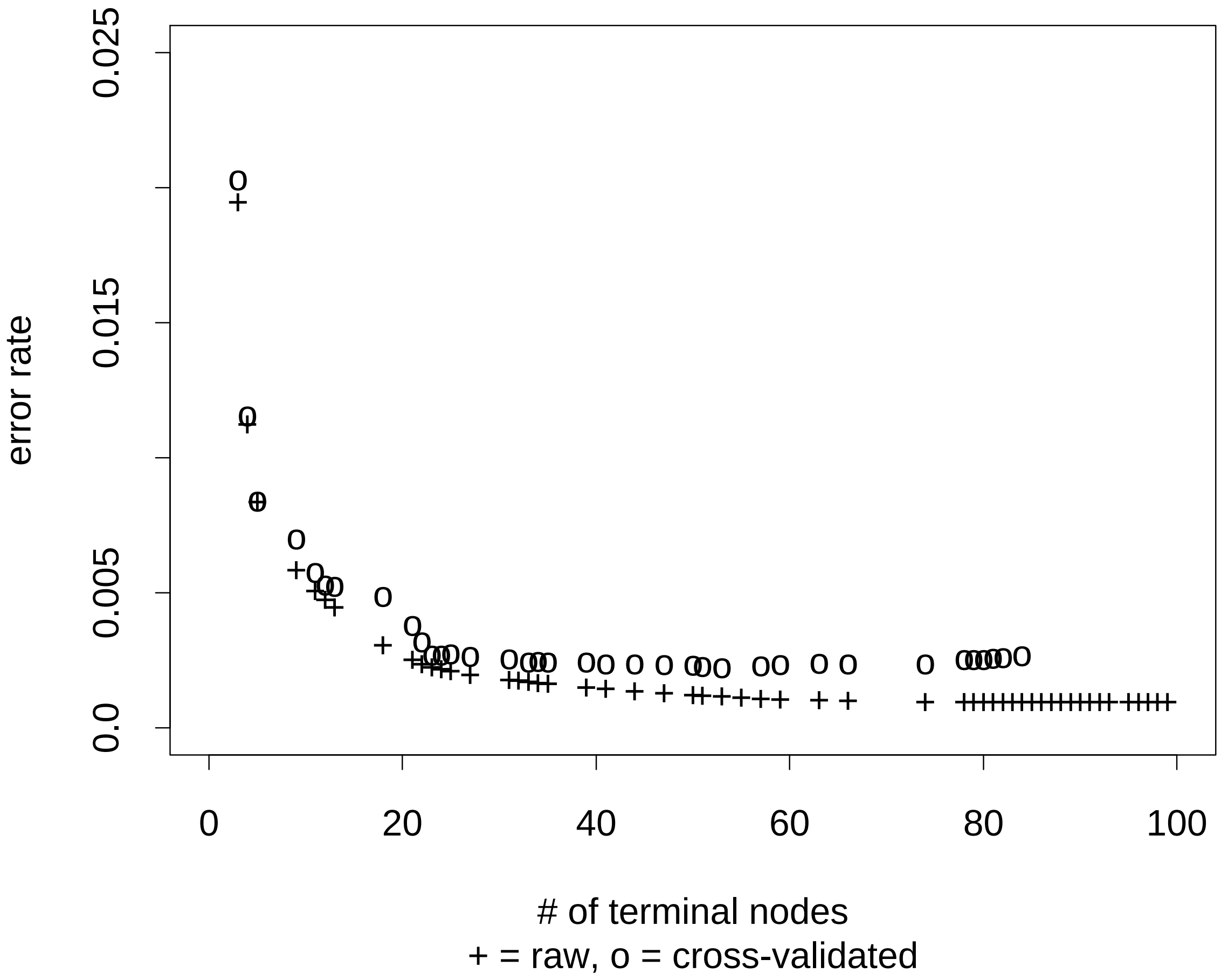

# 3. Decision Tree Node Assignment

Which class/value to assign to a terminal node?

- *Plurality vote*: Choose most frequent class at that node for classification; choose mean value for regression.

# End-of-Declarative-Sentence Prediction: Features

- Prob[word with "." occurs at end of sentence]
- Prob[word after "." occurs at beginning of sentence]
- Length of word with "."
- Length of word after "."
- Case of word with ".": Upper, Lower, Cap, Numbers
- Case of word after ".": Upper, Lower, Cap, Numbers
- Punctuation after "." (if any)
- Abbreviation class of word with ".": – e.g., month name, unit-of-measure, title, address name, etc.

# End of Declarative Sentence?

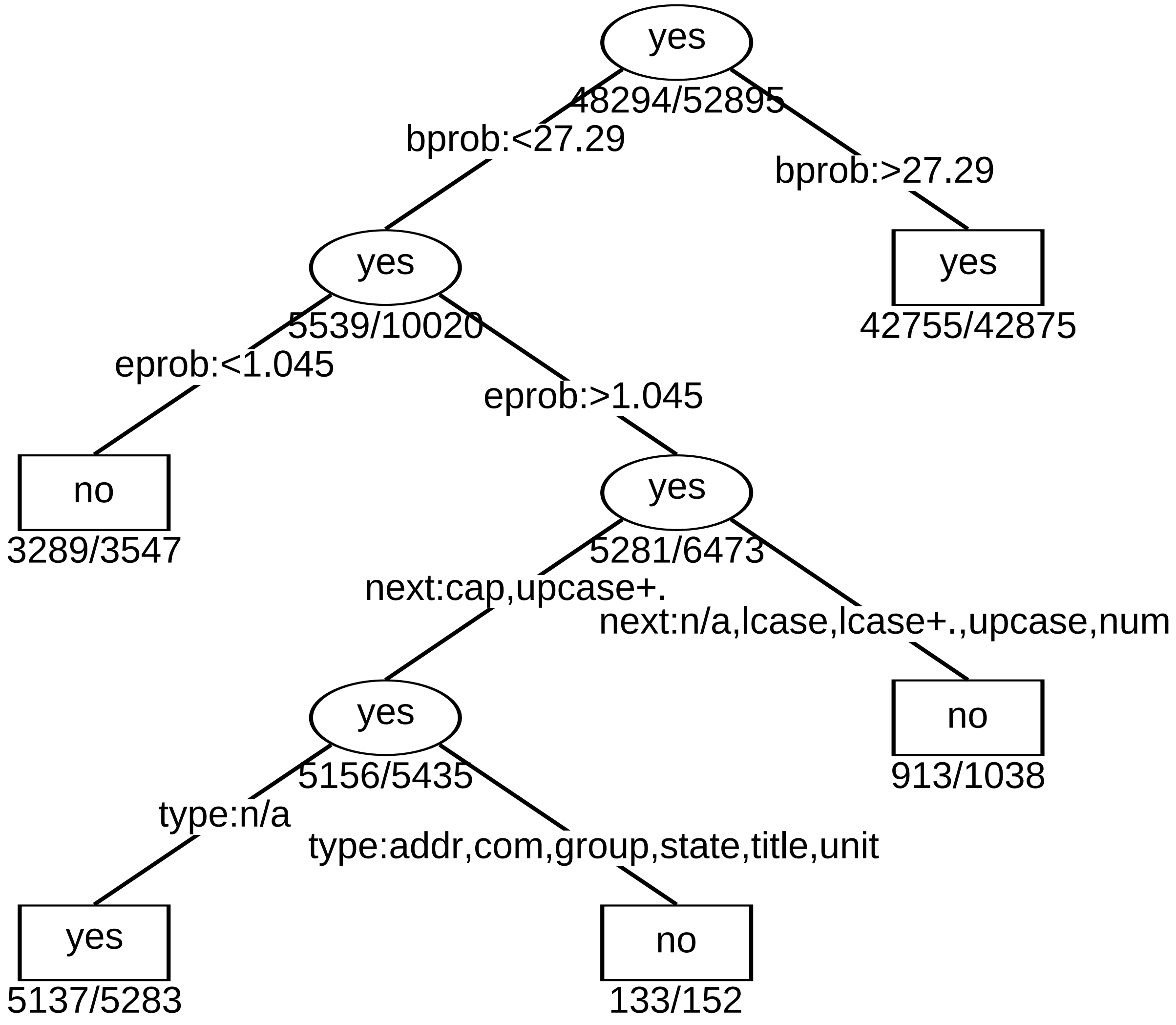

# Phoneme-to-Phone Alignment

| PHONEME | PHONE | WORD |
|---|---|---|
| p | p | purpose |
| er | er | |
| p | pcl | |
| - | p | |
| ax | ix | |
| s | s | |
| ae | ax | and |
| n | n | |
| d | - | |
| r | r | respect |
| ih | ix | |
| s | s | |
| p | pcl | |
| - | p | |
| eh | eh | |
| k | kcl | |
| t | t | |

# Phoneme-to-Phone Realization: Features

- Phonemic Context:
  - Phoneme to predict
  - Three phonemes to left
  - Three phonemes to right

- Stress (0, 1, 2)

- Lexical Position:
  - Phoneme count from start of word
  - Phoneme count from end of word

## Phoneme-to-Phone Realization: Prediction Example

Tree splits for `/t/` in ``` ``your pretty red'' ```:

| PHONE | COUNT | SPLIT |
|---|---|---|
| ix | 182499 | |
| n | 87283 | cm0: vstp,ustp,vfri,ufri,vaff,uaff,nas |
| kcl+k | 38942 | cm0: vstp,ustp,vaff,uaff |
| tcl+t | 21852 | cp0: alv,pal |
| tcl+t | 11928 | cm0: ustp |
| tcl+t | 5918 | vm1: mono,rvow,wdi,ydi |
| dx | 3639 | cm-1: ustp,rho,n/a |
| dx | 2454 | rstr: n/a,no |

(Riley, 1991).

# Phoneme-to-Phone Realization: Network Example

Phonetic network for `‘‘Don had your pretty...’’`:

| PHONEME | PHONE1 | PHONE2 | PHONE3 | CONTEXT |
|---|---|---|---|---|
| d | 0.91 d | | | |
| aa | 0.92 aa | | | |
| n | 0.98 n | | | |
| hh | 0.74 hh | 0.15 hv | | |
| ae | 0.73 ae | 0.19 eh | | |
| d | 0.51 dcl jh | 0.37 dcl d | | |
| y | 0.90 y | | | (if d→dcl d) |
| | 0.84 - | 0.16 y | | (if d→dcl jh) |
| uw | 0.48 axr | 0.29 er | | |
| r | 0.99 - | | | |
| p | 0.99 pcl p | | | |
| r | 0.99 r | | | |
| ih | 0.86 ih | | | |
| t | 0.73 dx | 0.11 tcl t | | |
| iy | 0.90 iy | | | |

# Decision Trees: Advantages/Limitations

- **Advantages:**
  - Handles continuous and categorical variables naturally.
  - Cross-validation gives results that generalize to new data.
  - Efficient algorithms – (approx. $nlogn$, $n =$ no. of obs.)
  - Small/medium-sized trees are easy interpret/modify.

- **Limitations:**
  - Recursive splitting can quickly cause “data starving” and replication of structure.
  - Categorical variables with large number of alternatives computationally unwieldy.
  - Large trees are hard to interpret/modify.

# The Replication Problem (*Pagallo and Haussler*)

- Smallest decision tree for DNF expression $x_1 x_2 + x_3 \overline{x}_4 x_5$:

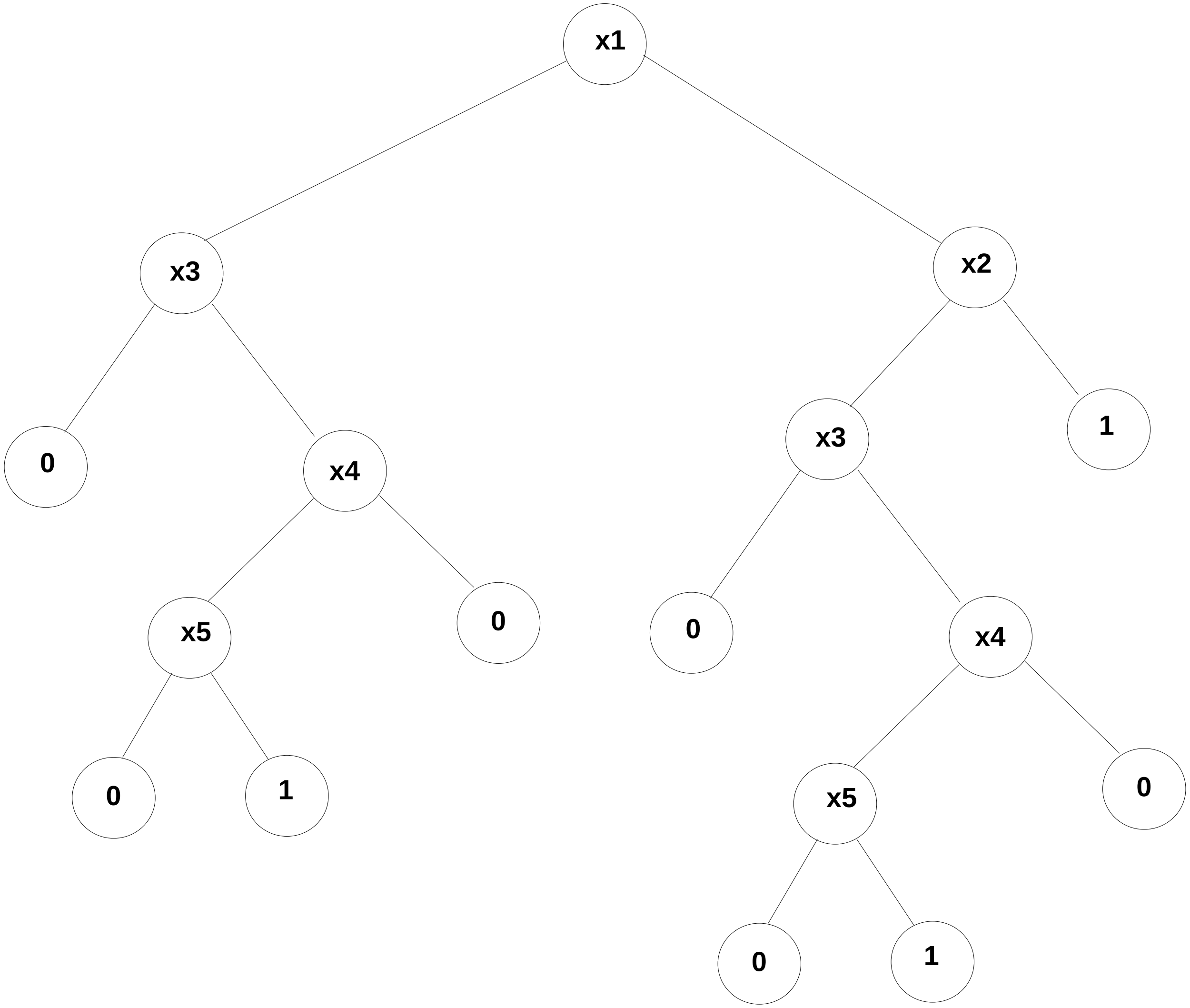

# The Replication Problem – Continued

- Equivalent tree using complex features in splits:

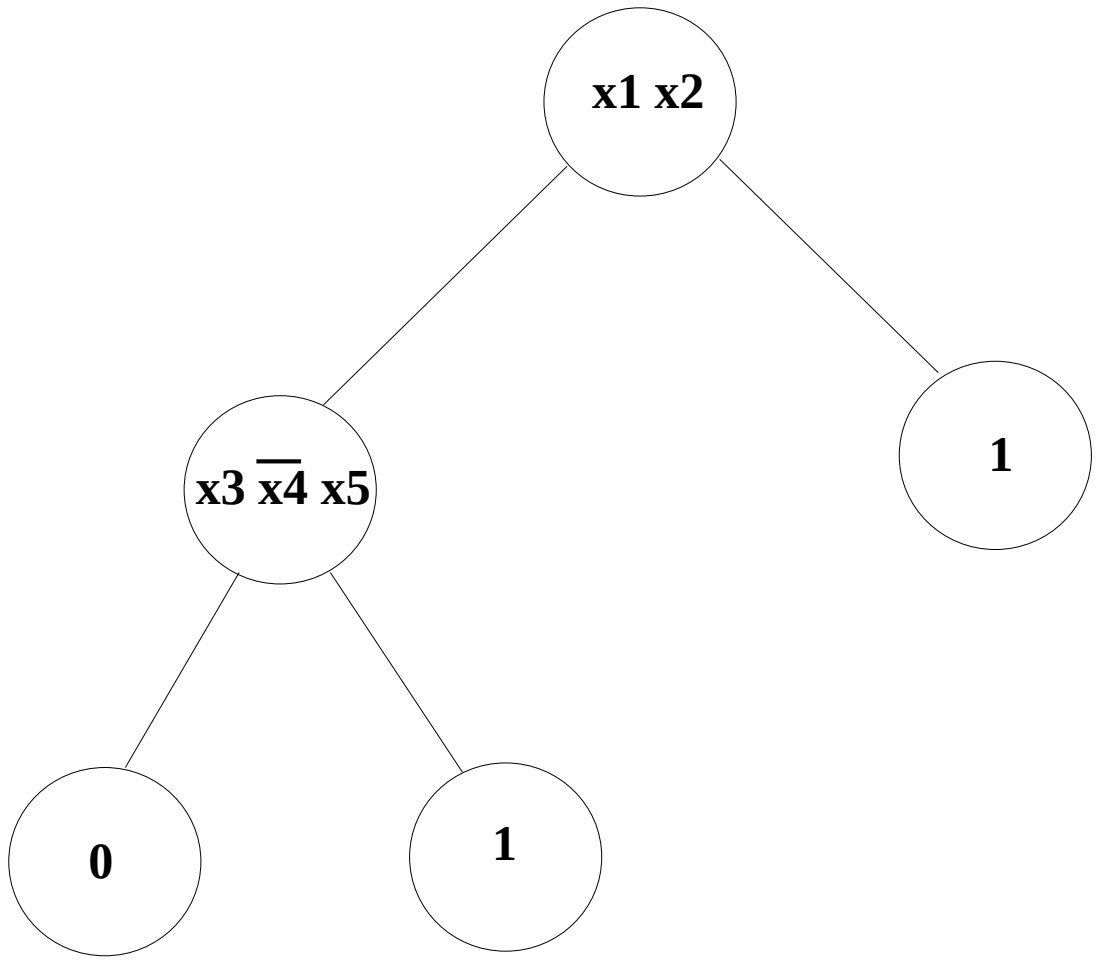

- Decision List Representation:

| Expression | Value |
|---|---|
| $x_1 x_2$ | 1 |
| $x_3 \overline{x}_4 x_5$ | 1 |
| *True* | 0 |

# Decision List Creation/Estimation

The *Separate and Conquer Algorithm* (*Pagallo and Haussler)* begins with a set of examples $S$, an auxilary set $P$ (*the pot*), and an empty decision list DL.

1. Select primitive feature that minimizes the entropy in $S$ after splitting on that feature.

2. Retain purer half of split in $S$ and place the rest in $P$. If there is only one class label in $S$, then goto 3, else goto 1.

3. Add complex feature that is the conjunction of the primitive features found in Steps 1 – 2 to the decision list DL with the class label of the examples in S.

4. If there is only one class label in $P$, then add that class label as the default case to the decision list DL and stop, else $S \leftarrow P$ and $P \leftarrow \Phi$ and goto 1.

## Decision List Creation/Estimation – Continued

- *Alternative approaches:*
  - Choose complex features from all combinations of $k$ primitive features (*Rivest*) or from particular combinations selected from the problem domain (*Yarowsky 1992, 1994*).
  - Do not partition the data after each decision, but reuse all the data, disallowing the previous decisions from recurring. More generally, interpolate by using a linear combination of the *global* and *residual* data (*Yarowsky 1992, 1994*).
- *Pruning Stategies:*
  - Evaluating held-out data, iteratively remove any decisions that do not improve the performance *(Pagallo and Haussler; Yarowsky 1992, 1994)*.
  - Remove any redundant decisions subsumed by prior decisions (Yarowsky, 1992, 1994).

# Lexical Ambiguity Resolution

- *Word sense disambiguation*:

  She handed down a harsh **sentence**. *peine*

  This **sentence** is ungrammatical. *phrase*

- *Homograph disambiguation*:

  He plays **bass**. /be$^{j}$s/

  This lake contains a lot of **bass**. /bæs/

- *Diacritic restoration*:

  appeler l'autre **cote** de l'atlantique côté 'side'

  **Cote** d'Azur côte 'coast'

(Yarowsky, 1992; Yarowsky 1994; Sproat, Hirschberg & Yarowsky, 1992; Hearst 1991)

# Homograph Disambiguation 1

- **N-Grams**

| Evidence | lɛd | lid | Logprob |
|---|---:|---:|---:|
| lead *level/N* | 219 | 0 | 11.10 |
| *of* lead *in* | 162 | 0 | 10.66 |
| *the* lead *in* | 0 | 301 | 10.59 |
| lead *poisoning* | 110 | 0 | 10.16 |
| lead *role* | 0 | 285 | 10.51 |
| *narrow* lead | 0 | 70 | 8.49 |
| lead *in* | 207 | 898 | 1.15 |

# Homograph Disambiguation 1

- **Predicate-Argument Relationships**

| | | | |
|---|---|---|---|
| *follow/V* + lead | 0 | 527 | 11.40 |
| *take/V* + lead | 1 | 665 | 7.76 |

- **Wide Context**

| | | | |
|---|---|---|---|
| *zinc* $\leftrightarrow$ lead | 235 | 0 | 11.20 |
| *copper* $\leftrightarrow$ lead | 130 | 0 | 10.35 |

- **Other Features (e.g. Capitalization)**

# Homograph Disambiguation 2

Sort by $Abs(Log(\frac{Pr(Pron_1|Collocation_i)}{Pr(Pron_2|Collocation_i)}))$

| Decision List for *lead* | | |
|---|---|---|
| Logprob | Evidence | Pronunciation |
| 11.40 | *follow/V* + lead | ⇒ lid |
| 11.20 | *zinc* ↔ lead | ⇒ lɛd |
| 11.10 | lead *level/N* | ⇒ lɛd |
| 10.66 | *of* lead *in* | ⇒ lɛd |
| 10.59 | *the* lead *in* | ⇒ lid |
| 10.51 | lead *role* | ⇒ lid |
| 10.35 | *copper* ↔ lead | ⇒ lɛd |
| 10.28 | lead *time* | ⇒ lid |
| 10.16 | lead *poisoning* | ⇒ lɛd |

# Homograph Disambiguation 3: Pruning

- **Redundancy by subsumption**

| Evidence | lid | lɛd | Logprob |
|---|---|---|---|
| lead *level/N* | 219 | 0 | 11.10 |
| lead *levels* | 167 | 0 | 10.66 |
| lead *level* | 52 | 0 | 8.93 |

- **Redundancy by association**

| Evidence | tɛɚ | tɪɚ |
|---|---|---|
| tear *gas* | 0 | 1671 |
| tear $\leftrightarrow$ *police* | 0 | 286 |
| tear $\leftrightarrow$ *riot* | 0 | 78 |
| tear $\leftrightarrow$ *protesters* | 0 | 71 |

# Homograph Disambiguation 4

Choose single best piece of matching evidence.

| **Decision List for *lead*** | | |
|---|---|---|
| Logprob | Evidence | Pronunciation |
| 11.40 | *follow/V* + lead | ⇒ lid |
| 11.20 | *zinc* ↔ lead | ⇒ lɛd |
| 11.10 | lead *level/N* | ⇒ lɛd |
| 10.66 | *of* lead *in* | ⇒ lɛd |
| 10.59 | *the* lead *in* | ⇒ lid |
| 10.51 | lead *role* | ⇒ lid |
| 10.35 | *copper* ↔ lead | ⇒ lɛd |
| 10.28 | lead *time* | ⇒ lid |
| 10.16 | lead *poisoning* | ⇒ lɛd |

# Homograph Disambiguation: Evaluation

| Word | Pron1 | Pron2 | Sample Size | Prior | Performance |
|---|---|---|---|---|---|
| lives | laɪvz | lɪvz | 33186 | .69 | .98 |
| wound | waʊnd | wund | 4483 | .55 | .98 |
| Nice | naɪs | nis | 573 | .56 | .94 |
| Begin | bɪ'gɪn | beɪgɪn | 1143 | .75 | .97 |
| Chi | tʃi | kaɪ | 1288 | .53 | .98 |
| Colon | koʊ'loʊn | 'koʊlən | 1984 | .69 | .98 |
| lead (N) | lid | lɛd | 12165 | .66 | .98 |
| tear (N) | tɛɚ | tɪɚ | 2271 | .88 | .97 |
| axes (N) | 'æksiz | 'æksɪz | 1344 | .72 | .96 |
| IV | aɪ vi | fɔɹθ | 1442 | .76 | .98 |
| Jan | dʒæn | jɑn | 1327 | .90 | .98 |
| routed | ɹutɪd | ɹaʊtɪd | 589 | .60 | .94 |
| bass | beɪs | bæs | 1865 | .57 | .99 |
| TOTAL | | | 63660 | .67 | .97 |

# Chinese Homograph Disambiguation

行 xing2/hang2

| CUE | PRON. | PRED. | | |
|---|---|---|---|---|
| 再 | xing2 | xing2 | 醫院 再 行 接洽 ， 解釋 | PROCEED, DO |
| 也 | xing2 | xing2 | 方案 也 行 不通 。 所以 | PROCEED, DO |
| 。 | xing2 | xing2 | 出來 。 行 ， 最 要緊 | PROCEED, DO |
| 孝 | xing2 | xing2 | 七 年 孝 行 模範 分別 | DEEDS |
| 梵 | xing2 | xing2 | 品 、 梵 行 品 。 有意 | DEEDS |
| 打字 | hang2 | hang2 | 球 打字 行 客戶 多 台 | COMPANY |
| 茶 | hang2 | hang2 | 性 到 茶 行 休息一下 | COMPANY |
| 央 | hang2 | hang2 | 行 。 央 行 總裁 張 繼 | COMPANY |
| 外 | hang2 | hang2 | 這 是 外 行 假充 內 行 | COMPANY |
| 百 | hang2 | hang2 | 有 一 百 行 或 更 大 | MEASURE WORD |
| 兩 | hang2 | hang2 | 望 ， 兩 行 淚水 晶晶 地 | MEASURE WORD |
| 船 | xing2 | xing2 | 風 ， 船 行 十分 ¬ 濡 | TRAVEL |
| 南 | xing2 | xing2 | ， 又 南 行 了 數 日 | TRAVEL |
| 路 | xing2 | xing2 | 直覺 得 行 路 有點 兒 | TRAVEL |
| 隨 | xing2 | xing2 | 傑 和 隨 行 代表 ， 表彰 | TRAVEL |
| | | | | |
| 是 | xing2 | hang2 | 是 由 同 行 的 人 暗中 | wrong |
| 者 | hang2 | xing2 | 幾 名 同 行 的 業 者 | wrong |

************************************************* PERCENT CORRECT: 94

# Decision Lists: Advantages/Limitations

- **Advantages:**
  - Efficient and flexible use of data.
  - Easy to interpret and modify.
  - Handles both wide and narrow context information.

- **Limitations:**
  - New area; many aspects not well-studied – e.g., best complex feature selection rules, efficient pruning/cross-validation techniques, global vs. residual vs. interpolated dataset division.

## Mixed Methods: Trained Hand-crafted Rules

- Probabilistic parsing (e.g. Jelinek et al., 1990; Su et al., 1992; Goddeau, 1992)

- Probabilistic morphological analysis (e.g. Heemskerk, 1993)

  Consider that the Dutch word *beneveling* 'intoxication' has two morphologically possible analyses (the second depends upon regular Dutch orthographic spelling changes):

| | | |
|---|---|---|
| $be_{V/N}+nevel_N+ing_{V\backslash N}$ | BE- + mist + -ion | 'intoxication' |
| $be_{V/N}+neef_N+eling_{V\backslash N}$ | BE- + nephew + -ling | '??' |

| | |
|---|---|
| Prefixation: | $B/A \cdot A \rightarrow B$ |
| Suffixation: | $A \cdot A\backslash B \rightarrow B$ |
| Compounding: | $A \cdot B \rightarrow B$ |

# Probabilistic Morphological Parsing

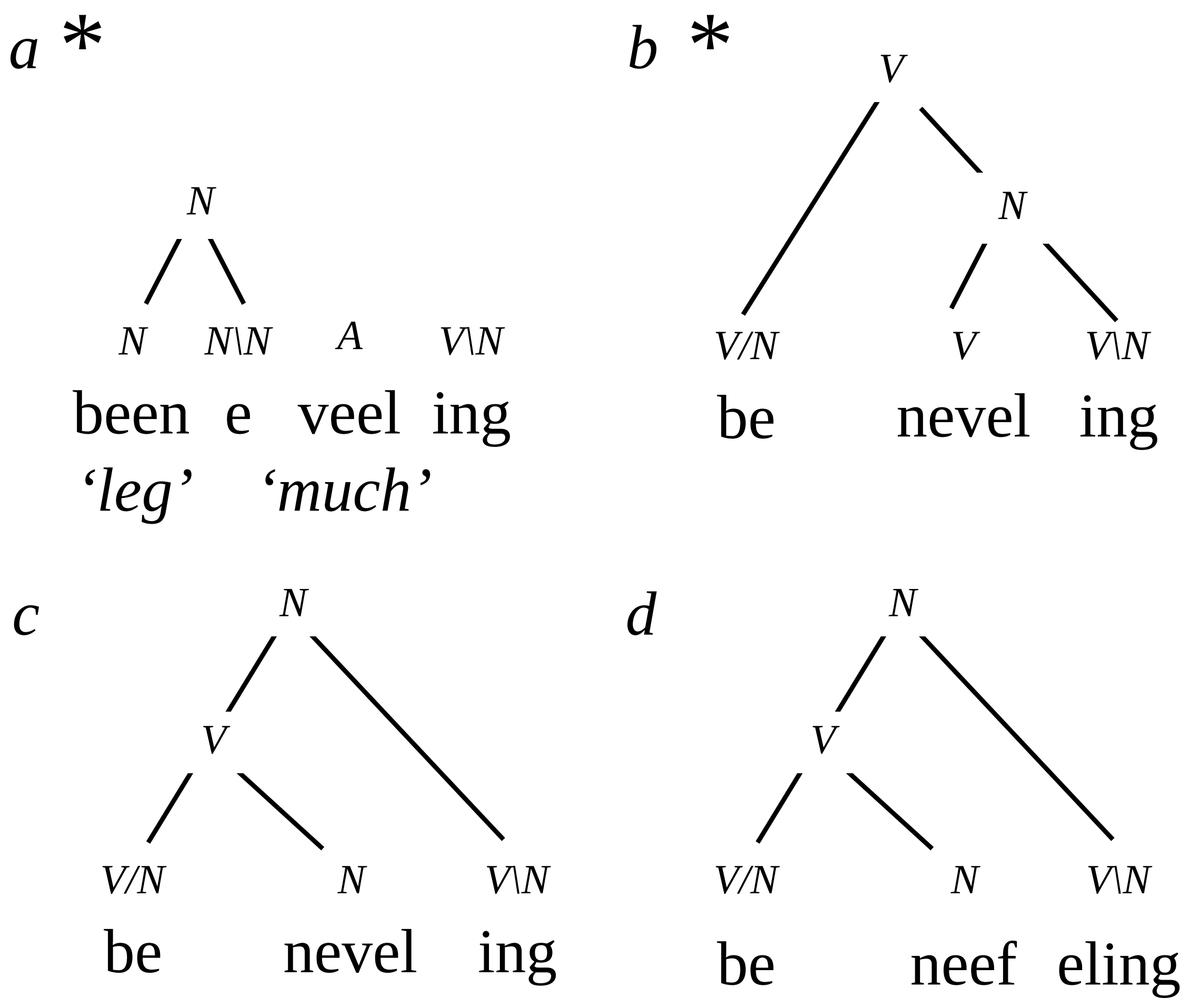

(a) is ruled out by (hand-constructed) categorial grammar, while
(b) is ruled out by prohibiting noun morphology from being input to verb morphology

# Probabilistic Morphological Parsing

$$p([_N[_V[_{V/N}\textbf{be}][_N\textbf{nevel}]][_{V\backslash N}\textbf{ing}]]) =$$
$$p(word{\rightarrow}N)\times$$
$$p(N{\rightarrow}V\ V\backslash N)\times$$
$$p(V{\rightarrow}V/N\ N)\times$$
$$p(V/N{\rightarrow}\textbf{be})\times$$
$$p(N{\rightarrow}\textbf{nevel})\times$$
$$p(V\backslash N{\rightarrow}\textbf{ing})$$

$$p([_N[_V[_{V/N}\textbf{be}][_N\textbf{nevel}]][_{V\backslash N}\textbf{ing}]]) >$$
$$p([_N[_V[_{V/N}\textbf{be}][_N\textbf{neef}]][_{V\backslash N}\textbf{eling}]])$$

# Probabilistic Morphological Parsing

- Probabilities for production rules were calculated from the CELEX database containing 123,000 morphologically annotated Dutch stems.

- Among 1612 structurally ambiguous words that had a correct analysis among the alternatives, probabilistic techniques gave the highest score to the correct analysis in $1483/1612 = \mathbf{92}$% of the cases.

# Weighted Finite-State Methods: Motivation

- Unified representation for information sources:
  - strings/lattices
  - dictionaries
  - decoders/generators
  - language models
  - . . .
- Uniform algorithms for:
  - combining information sources into generators, decoders, etc.
  - search
  - minimizing representations
- Modular definition of language processors (cf. *lex*, *yacc*)

# Weighted Finite-State Methods: Origins

- Probabilistic automata (Paz, Taylor and Booth,. . . )
- Algebraic theory of languages and automata (Schutzenberger, Eilenberg, Berstel, Kuich & Salomaa,. . . )
- Hidden Markov models (. . . )
- Theory of shortest path algorithms (Dijkstra, Aho, Hopcroft &Ullman,. . . )

# Transduction Cascades

- Standard "noisy channel" model: for given observations $o$, find message $w$ that maximizes

$$P(w,o) = P(o|w)P(w)$$

- Multistage cascade:

$$P(s_0, s_k) = P(s_k|s_0)P(s_0)$$
$$P(s_k|s_0) = \sum_{s_1,\ldots,s_{k-1}} P(s_k|s_{k-1}) \cdots P(s_1|s_0)$$

- "Viterbi" version:

$$\tilde{P}(s_0, s_k) \quad = \quad \tilde{P}(s_k|s_0) + \tilde{P}(s_0)$$
$$\tilde{P}(s_k|s_0) \quad \approx \quad \min_{s_1,\ldots,s_{k-1}} \sum_{1 \le j \le k} \tilde{P}(s_j|s_{j-1})$$

where $\tilde{X} = -\log X$.

# The Basic Generalizations

- *Weighted languages:* functions from strings to weights, modeling information sources
- *Weighted transductions:* functions from pairs of strings (one from each level) to weights, modeling mappings between levels of representation
- *Rational algebra:* make complex languages and transductions from simple ones
- **Examples:**
  - *languages:* phone sequence/lattice, language model
  - *transductions:* pronunciation dictionary, phoneme-to-phone realization, grapheme-to-phoneme conversion, text segmenter

# Weights

- *Weight semiring:* set of weights $K$ with two commutative, associative operations:
  - *sum:* combines the weights of the ways of deriving an object to form the overall weight of the object
  - *product:* combines the weights of subobjects into the weight of the combined object;
  - product distributes with respect to sum
- **Examples:**

| semiring | sum | product | 0 | 1 |
|---|---|---|---|---|
| Viterbi | min | $+$ | $+\infty$ | 0 |
| probability | $+$ | $\times$ | 0 | 1 |
| boolean | or | and | 0 | 1 |

# Weighted Languages and Transductions

- Generalized information source: *weighted language*

$$L: \quad \Sigma^* \quad \rightarrow \quad K$$

behaviors    weights

- Generalized transduction step: *weighted transduction*

$$S: \quad \Sigma^* \quad \times \quad \Gamma^* \quad \rightarrow \quad K$$

inputs    outputs    weights

- Combining levels — *generalized composition*:

*composition:* $(S \circ T)(r,t) = \sum_{s \in \Gamma^*} S(r,s)T(s,t)$

*application:* $(L \circ S)(s) = \sum_{r \in \Sigma^*} L(r)S(r,s)$

*reverse application:* $(S \circ M)(r) = \sum_{s \in \Gamma^*} S(r,s)M(s)$

*intersection:* $(M \circ N)(s) = M(s)N(s)$

# Rational Operations

- Making complex languages and transductions from simple ones:

| | |
|---|---|
| singleton | $\{\mathbf{u}\}(\mathbf{v}) = 1$ iff $\mathbf{u} = \mathbf{v}$ |
| scaling | $(kX)(\mathbf{u}) = kX(\mathbf{u})$ |
| sum | $(X + Y)(\mathbf{u}) = X(\mathbf{u}) + Y(\mathbf{u})$ |
| concatenation | $(XY)(\mathbf{w}) = \sum_{\mathbf{u}\cdot\mathbf{v}=\mathbf{w}} X(\mathbf{u})Y(\mathbf{v})$ |
| power | $X^0(\boldsymbol{\epsilon}) = 1, X^0(\mathbf{u} \neq \boldsymbol{\epsilon}) = 0, X^{n+1} = XX^n$ |
| closure | $X^* = \sum_{k \geq 0} X^k$ |

- Example— pronunciation dictionary:

| | |
|---|---|
| $D_w(p, w)$ | probability that word $w$ is realized as phone string $p$ |
| $\left(\sum_w D_w\right)^*$ | context-independent probabilities for realizations of word strings as phone strings |

# Weighted Automata

Weighted finite automata implement rational weighted languages and transductions

- Automata transitions:

$$q \stackrel{\mathbf{x}/k}{\longrightarrow} q'$$

acceptor: $\mathbf{x} \in \Sigma \cup \{\epsilon\}$; transducer: $\mathbf{x} \in (\Sigma \cup \{\epsilon\}) \times (\Gamma \cup \{\epsilon\})$; weight $k$

- Example — word pronunciation transducer:

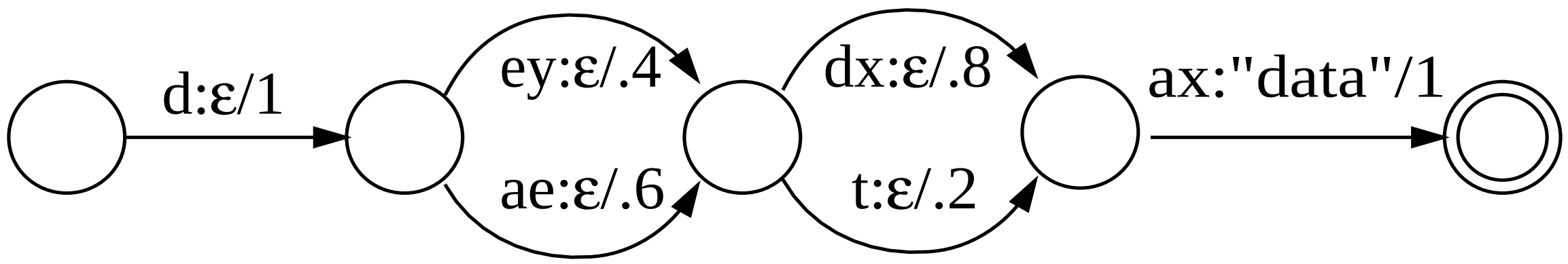

# Automata Operations

- Operations between weighted rational languages and transductions have corresponding automata operations
- Generalized composition is implemented by a general *automata join* operation

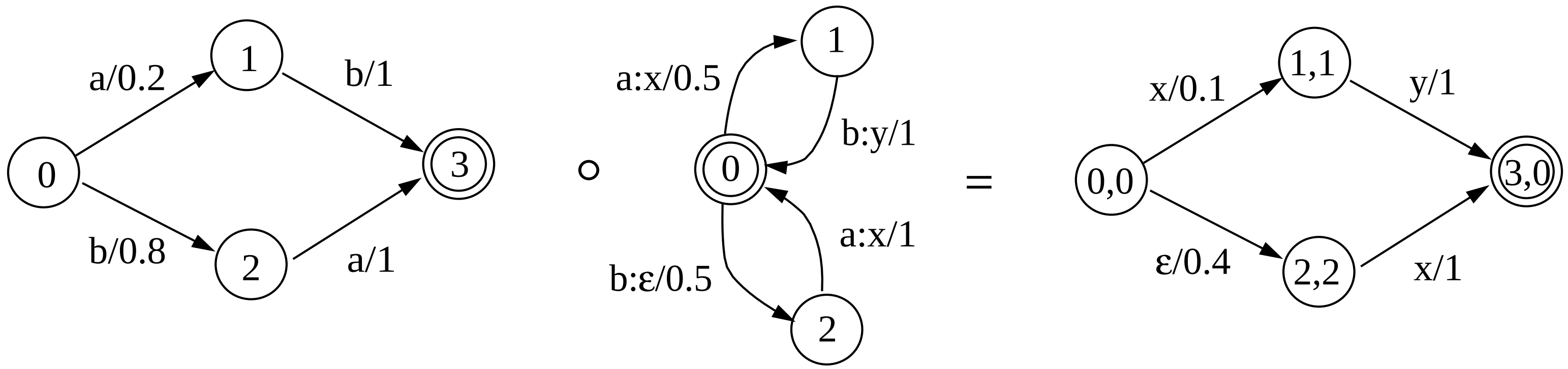

- *Pruning:* keep only those paths in the join within a beam of the best path
- *Optimization:* try to find a smaller weighted automaton with the same language/transduction

# Rule Compilation

- Theory of phonological rewrite rules and their implementation as finite-state transducers is well understood (Kaplan & Kay, 1994)
- E.g.: (left-to-right) obligatory rule of the form

$$\phi \rightarrow \psi / \lambda \underline{\quad\quad} \rho$$

can be modeled by composing a series of transducers

$$\begin{gathered} Prologue \circ \\ Id(Obligatory(\phi, <_i, >)) \circ \\ Id(Rightcontext(\rho, <, >)) \circ \\ Replace \circ \\ Id(Leftcontext(\lambda, <, >)) \circ \\ Prologue^{-1} \end{gathered}$$

each of which expresses a regular relation that restricts a certain portion of the application of the rule.

# Rule Compilation: Statistical Rules

A sample (toy) probabilistic ruleset

```
{All} := ptkdaeiou\&R012 ;;
{Cons} := ptkd ;;
{Vowel} := aeiou\&R ;;
{Stress} := 12 ;;

End Prolog

t -> ({DD}<0.20>, {tt}<4.32>, {dd}<4.64>,
      {??}<5.64>, {DEL}<6.64>)
        / {Stress}{Cons}*{Vowel} __ 0{Vowel} ;;
k -> ({??}<0.15>, {kk}<3.32>) / __ # ;;
R -> ({&&}<0.15>, {RR}<0.14>) / 0 __ ;;
t -> ({tt}<4.32>, {??}<1.8>) / __ # ;;
```

# Rule Compilation: Statistical Rules

Output for *1at0Rk* given compiled ruleset from previous slide

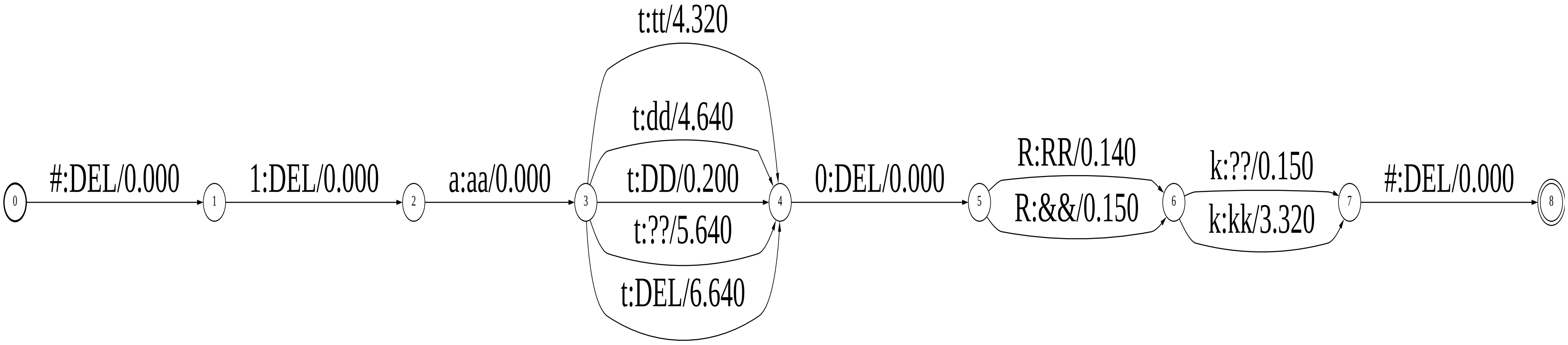

# Chinese Word Segmentation

| I | forget | NEG−POT | liberation | avenue | be−at | where |
|---|---|---|---|---|---|---|
| 我 | 忘 | 不 了 | 解 放 | 大 街 | 在 | 哪 裡 |

(understand) (enlarge)

“I couldn’t forget where Liberation Avenue is.”

# Chinese Word Segmentation

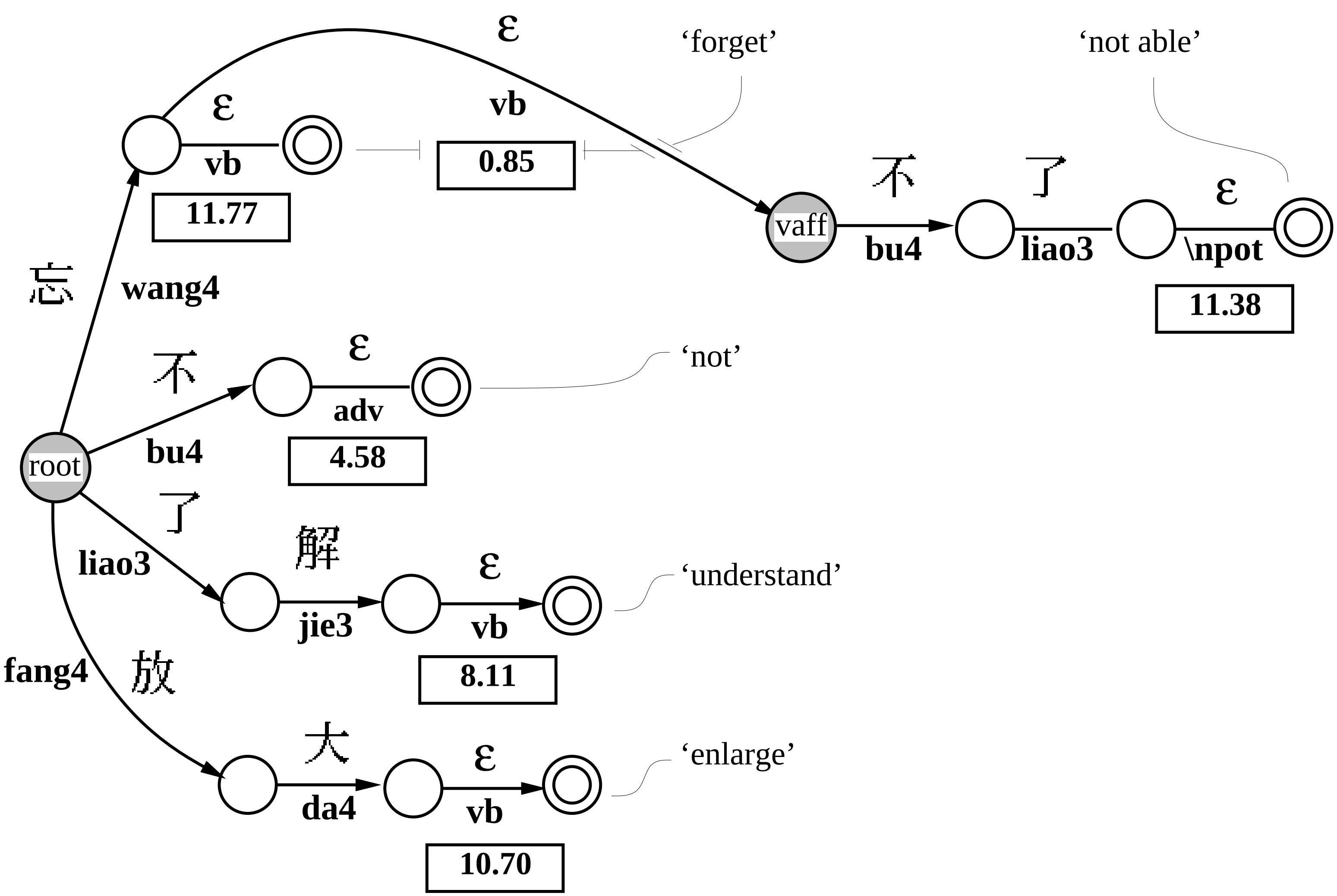

# Chinese Word Segmentation

"I couldn't forget where Liberation Avenue is."

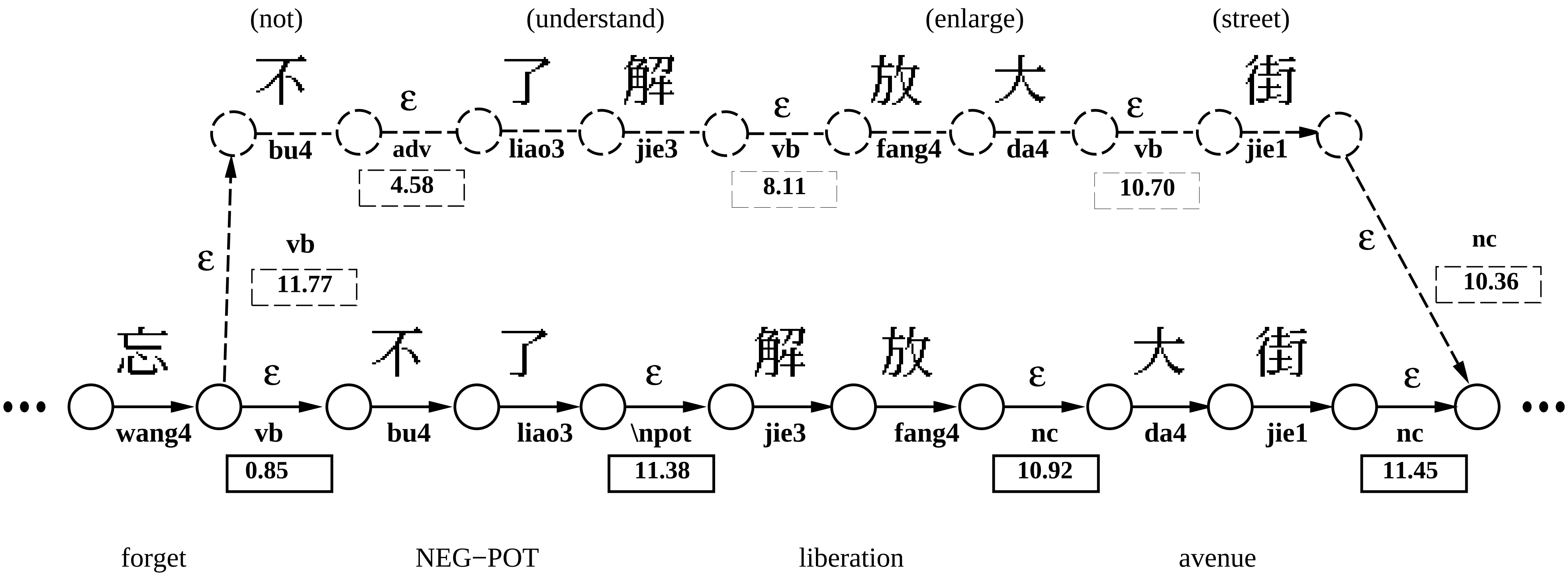

$$0.85 + 11.38 + 10.92 + 11.45 = 34.60$$

wins over

$$11.77 + 4.58 + 8.11 + 10.70 + 10.36 = 45.52$$

## Some Text-Analysis Problems in Text-to-Speech Synthesis

- Text-normalization issues
  - End-of-sentence detection
  - Word-segmentation (Chinese, Japanese, Thai)
  - Abbreviation expansion: is **St.** *Saint* or *Street*?
  - Numeral interpretation: is **747** *seven hundred and forty seven*, or *seven forty seven*?
- Part-of-speech assignment
- Word pronunciation
  - Morphological analysis of ordinary words and names
  - Homograph disambiguation
- Accent prediction
- Prosodic phrasing prediction

# Pitch Accent Prediction

**Problem**: predict accent status for different classes of words

- Function versus content word:
  JOHN GAVE it to <u>BILL</u>
  He GAVE it to <u>him</u>

- Long noun phrases:
  CITY <u>HALL</u>
  TAX <u>office</u>
  CITY <u>hall</u> TAX office

- Preposing:
  We will BEGIN to LOOK at FROG anatomy <u>today</u>
  <u>TODAY</u> we will BEGIN to LOOK at FROG anatomy

- Information status:
  My SON WANTS me to BUY a <u>DOG</u>, but I’m ALLERGIC to <u>dogs</u>.

# Pitch Accent Prediction: Sample Variables

For each word $w$:

- distance of $w$ from beginning/end of sentence
- total words in utterance
- distance of $w$ in words from/to prior/next boundary
- part-of-speech of $w$
- if $w$ is in complex nominal, the predicted accent of $w$ given by an automatic noun-phrase accent predictor (Sproat, 1994)
- information status

# Pitch Accent Prediction: Sample Tree

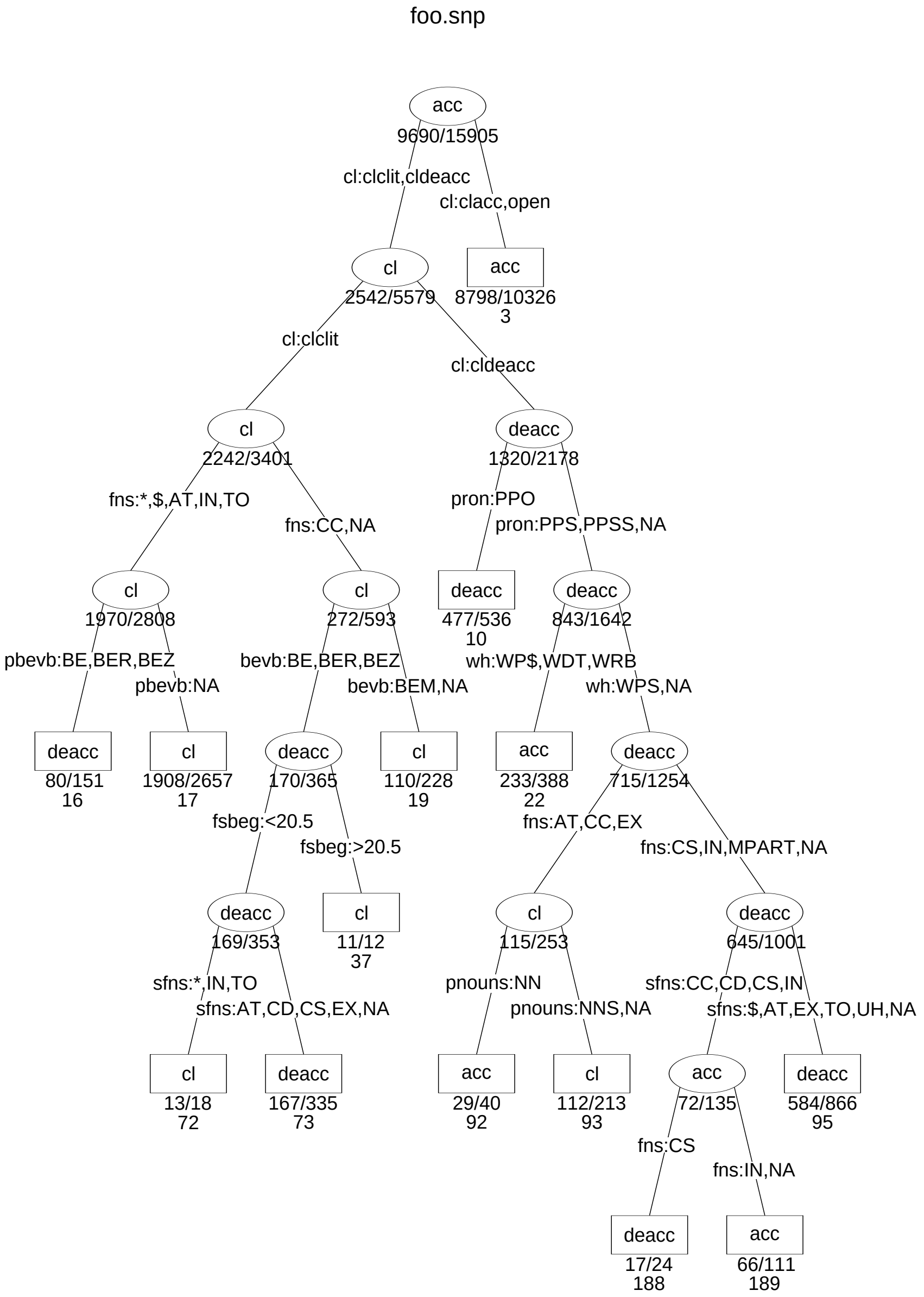

# Pitch Accent Prediction: Results

- Pitch Accents Predicted from the Audix Corpus: **80% Correct**
- Pitch Accents Predicted from the ATIS Database: **81.9% Correct**
- ATIS Predictions with Boundary Information: **85.1% Correct**
- ATIS Predictions with Speaker Information: **85.1% Correct**

# Pitch Accent Prediction: Hand-derived decision rules

*For each item $w_i$ labeled with part-of-speech $p_i$:*

If $w_i$ is a phrasal verb, deaccent;

Else if $p_i$ is classified 'closed-cliticized', cliticize;

Else if $p_i$ is classified 'closed-deaccented', deaccent;

Else if $w_i$ is marked 'contrastive', 'prefixed', or 'preposed', assign it emphatic accent;

Else if $w_i$ is part of a proper nominal

> If $w_i$'s status is 'given', assign emphatic accent,
> else assign a simple pitch accent;

Else if $w_i$ is in global focus but not in local focus ('given'), assign emphatic accent;

Else if $w_i$ is classified 'closed-accented', accent;

Else if $w_i$ is in local focus ('given'), deaccent;

Else if $w_i$ is part of a (common) complex nominal

> If $w_i$ is predicted to be accented in citation form, accent
> else deaccent;

Else accent $w_i$.

# Word Pronunciation

- Represent lexicon as FSA $L$.
  - Morphological derivatives of words in the lexicon can be represented using standard finite-state morphological techniques (Koskenniemi, 1983; Karttunen et al. 1992; Sproat 1992).
  - Corpus-derived weights can be added to arcs to rank multiple analyses.
  - Orthographic rules can be compiled into a (W)FST $O$ that can be composed with the lexicon $L$ to form a (W)FST $L'$ that can morphologically decompose words as they occur in text.
  - Pronunciation rules (either hand-built or compiled from a trained decision tree) can be compiled into a (W)FST $P$ that can be composed with $L'$ to yield a (W)FST $L''$ that will transduce input words to sets of pronunciations.
- $L''$ can be composed with a WFST $\Phi$ implementing phoneme-to-phone rules (again, either hand-developed or compiled from a trained decision tree) to yield $L''' = L'' \circ \Phi$. $L'''$ can then be inverted for use in an ASR system.

# Word Pronunciation: an Example

| | | |
|---|---|---|
| 1 | **Input** | отцов |
| 2 | **Annotated Input** | отц"ов |
| 3 | **Underlying Form** | от"{E1}ц{noun}{msc}{an}+"ов{pl}{gen} |
| 4 | **Phonological Form** | ats"of |

- Map 1 to 2 by transducer that freely introduces stress marks (").
- Compose lexicon of legal underlying forms, with rules such as

  $\{E1\} \rightarrow$ e / __ $\{DelCons\}\{GRAM\}^{*}+\{NUM\}$

  $\{E1\} \rightarrow \emptyset$

  " $\rightarrow \emptyset$/ $_\Sigma^{*}$ "

  Invert the resulting transducer and compose this with 2 to produce 3.
- Compile pronunciation rules such as

  o $\rightarrow$ o / "__

  o $\rightarrow$ a

  into transducer that can be composed with 2 to produce 4.

# Phrasing Prediction

- **Problem**: predict intonational phrase boundaries in long unpunctuated utterences:

  *For his part, Clinton told reporters in Little Rock, Ark., on Wednesday || that the pact can be a good thing for America || if we change our economic policy || to rebuild American industry here at home || and if we get the kind of guarantees we need on environmental and labor standards in Mexico || and a real plan || to help the people who will be dislocated by it.*

- Previous treatments have used rule-based parsing approaches (O'Shaughnessy, 1989; Bachenko & Fitzpatrick, 1990).
- AT&T synthesizer uses a CART-based predictor trained on labeled corpora (Hirschberg & Wang 1992).

# Phrasing Prediction: Variables

For each $< w_i, w_j >$:

- length of utterance; distance of $w_i$ in syllables/ stressed syllables/words ... from the beginning/end of the sentence
- automatically predicted pitch accent for $w_i$ and $w_j$
- part-of-speech (POS) for a 4-word window around $< w_i, w_j >$;
- (largest syntactic constituent dominating $w_i$ but not $w_j$ and vice versa, and smallest constituent dominating them both)
- whether $< w_i, w_j >$ is dominated by an NP and, if so, distance of $w_i$ from the beginning of that NP, the NP, and distance/length
- (mutual information scores for a four-word window around $< w_i, w_j >$)

The most successful of these predictors so far appear to be POS, some constituency information, and mutual information

# Phrasing Prediction: Sample Tree

new.snp

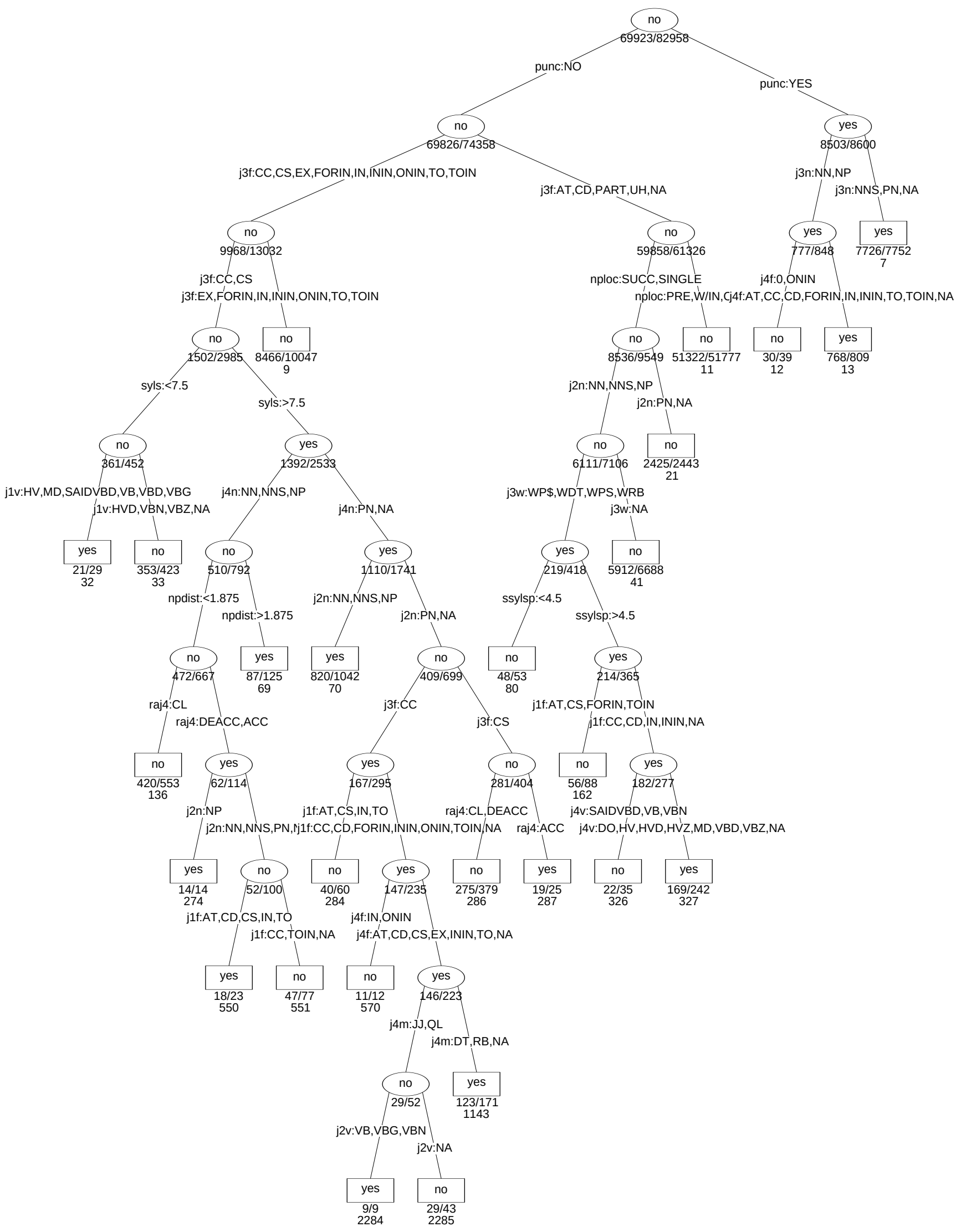

# Phrasing Prediction: Results

- Results for multi-speaker read speech:
  - major boundaries only: **91.2%**
  - collapsed major/minor phrases: **88.4%**
  - 3-way distinction between major, minor and null boundary: **81.9%**
- Results for spontaneous speech:
  - major boundaries only: **88.2%**
  - collapsed major/minor phrases: **84.4%**
  - 3-way distinction between major, minor and null boundary: **78.9%**
- Results for 85K words of hand-annotated text, cross-validated on training data: **95.4%**.

# AT&T Text-to-Speech Synthesis

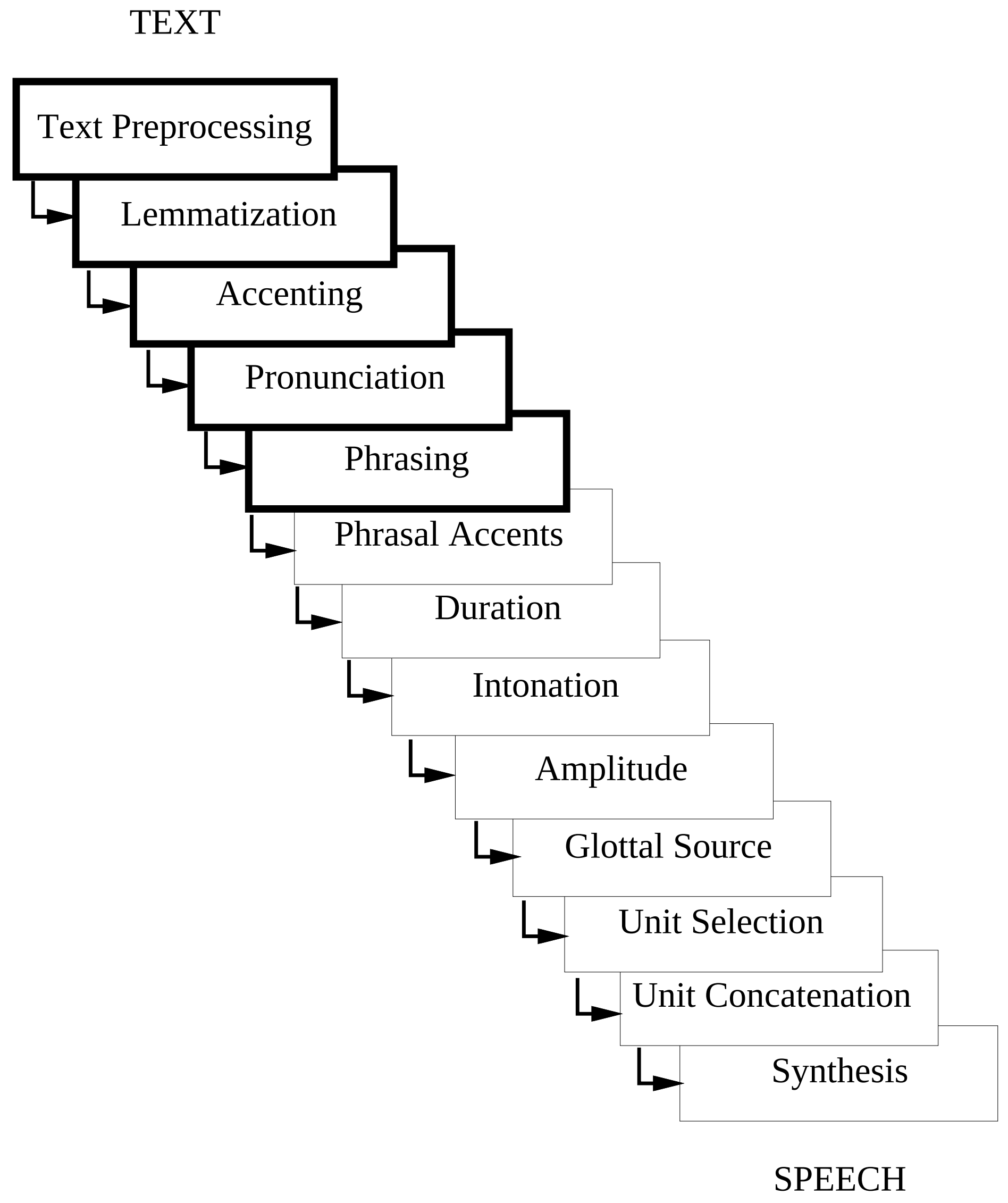

# Representations in Speech Recognition

- Quantized observations:

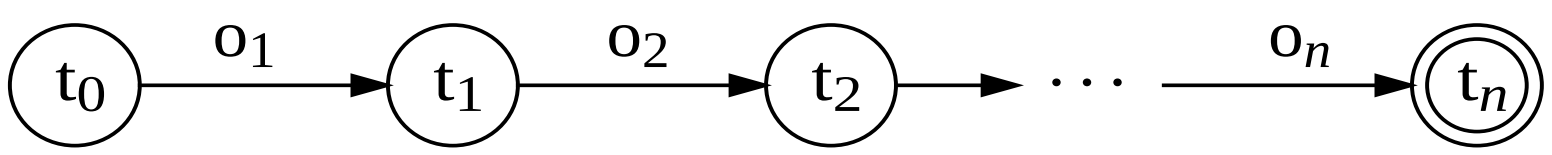

- Phone model $A_\pi$:

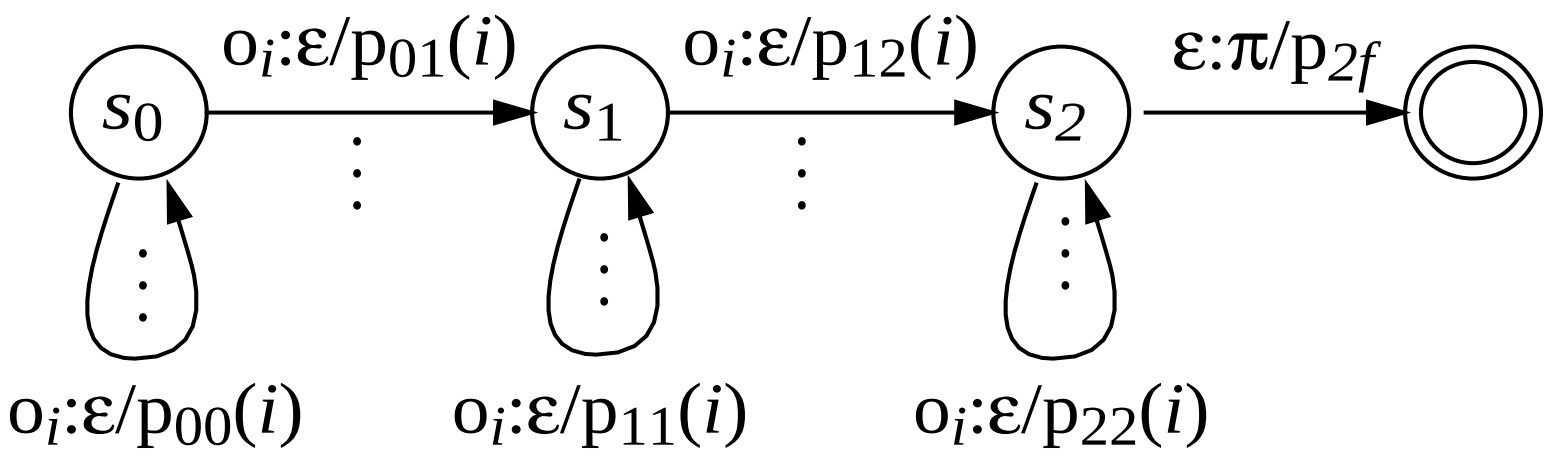

Acoustic transducer: $A = \left(\sum_\pi A_\pi\right)^*$

- Word pronunciations $D_{\text{data}}$:

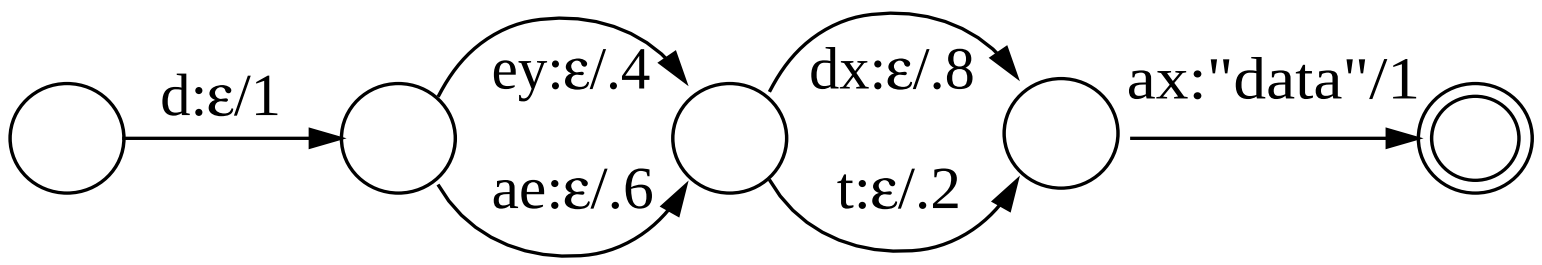

Dictionary: $D = \left(\sum_w D_w\right)^*$

# Recognition Cascade

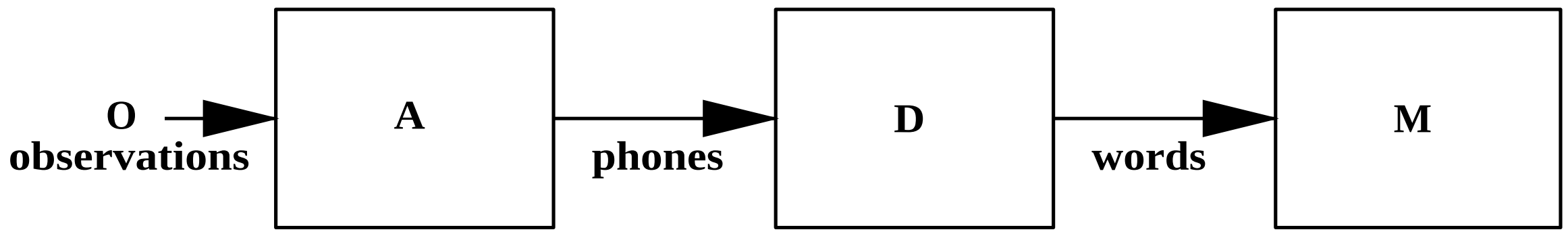

- Levels:
  - *Observations:* $O(o) = 1, O(s \neq o) = 0$
  - *Acoustic-phone transduction:* $A(a, p) = P(a|p)$
  - *Pronunciation dictionary:* $D(p, w) = P(p|w)$
  - *Language model:* $M(w) = P(w)$
- Recognition: maximize $(O \circ A \circ D \circ M)(w)$

# Example: Phone Lattice $O \circ A$

Phone lattice for *hostile battle*:

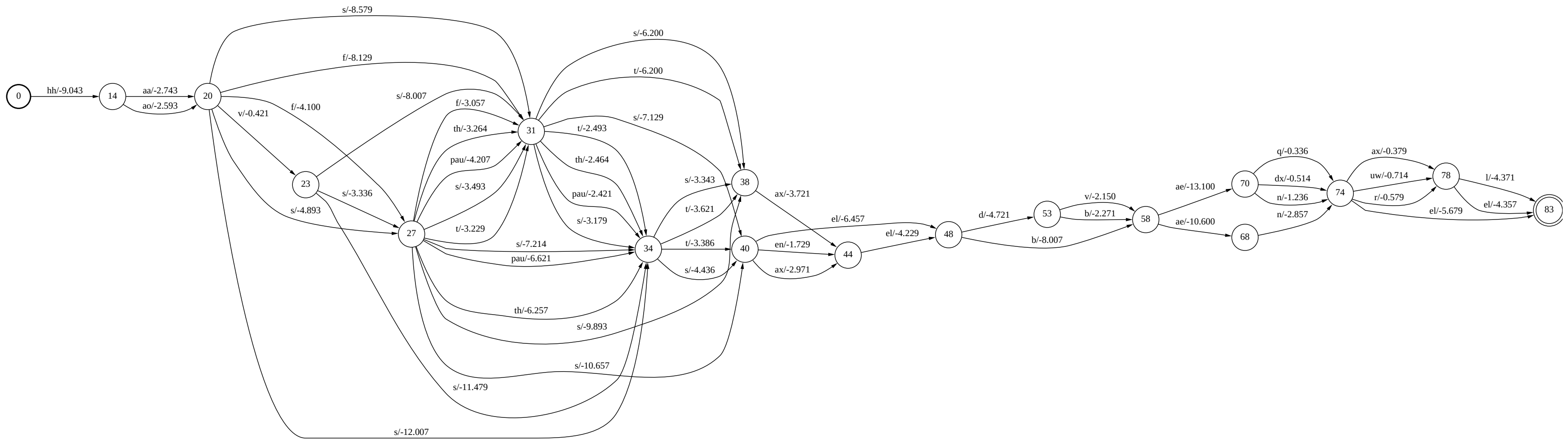

# Sample Pronunciation Dictionary $D$

Dictionary with *hostile*, *battle* and *bottle* as a transducer:

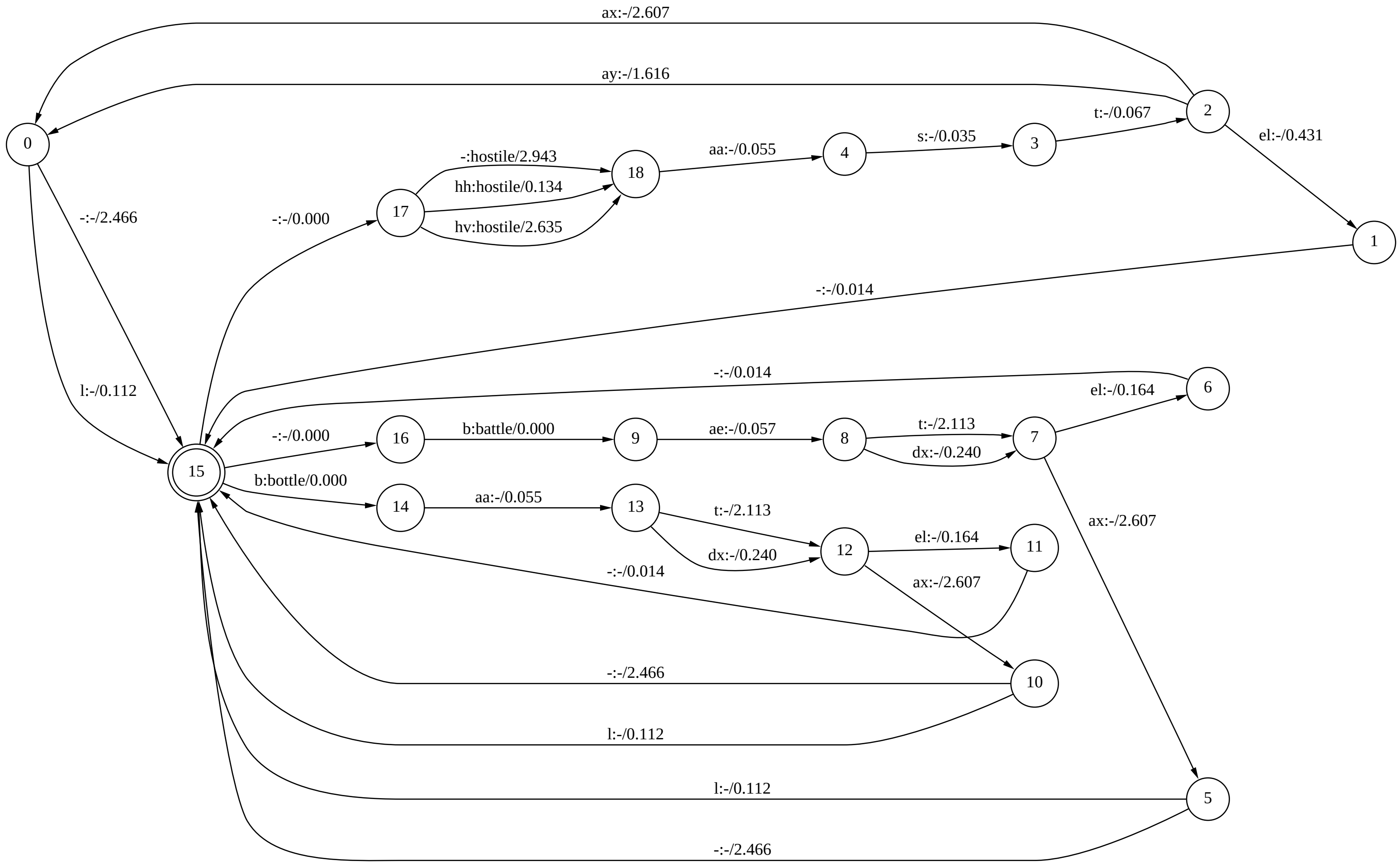

# Sample Language Model $M$

Language model as acceptor:

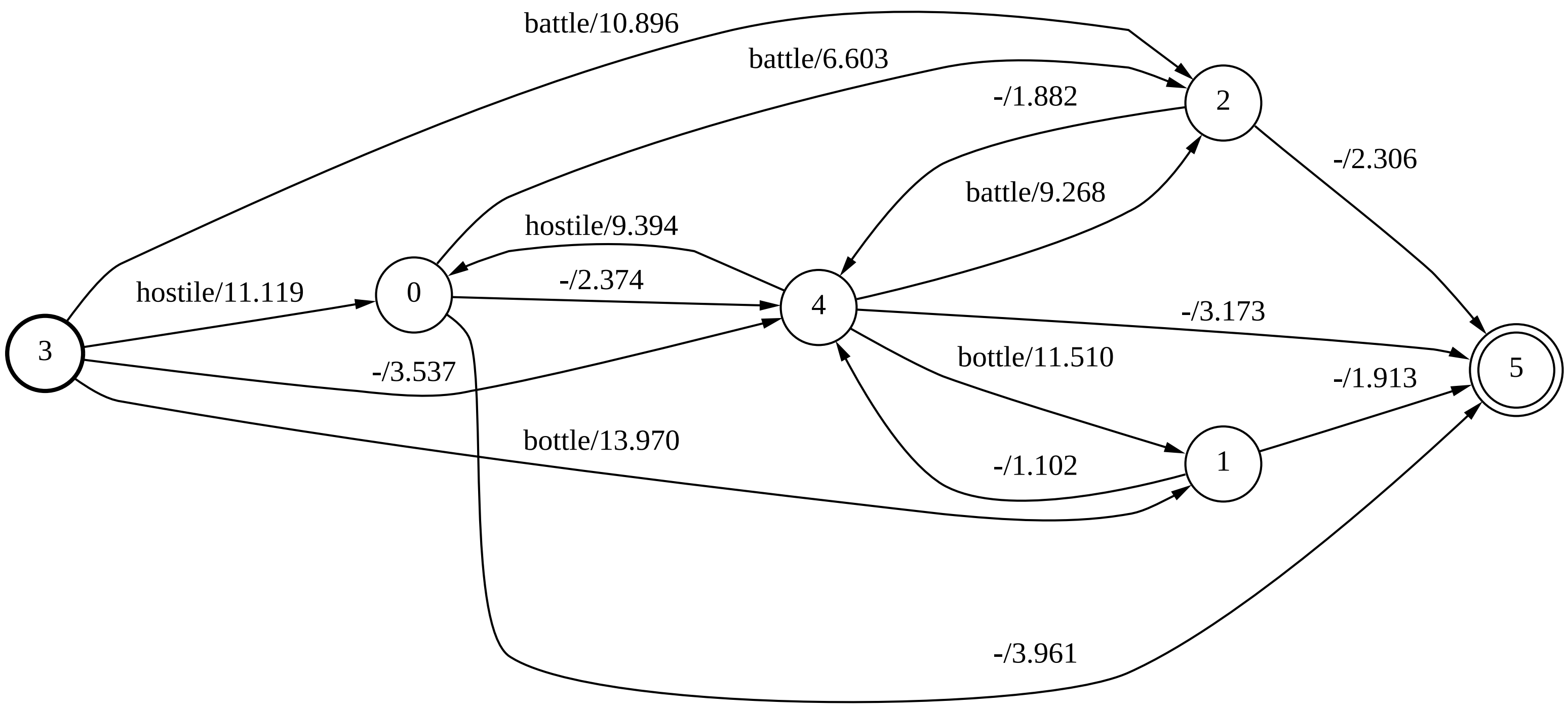

# Recognition Output: $O \circ A \circ D \circ M$

Apply dictionary to phone lattice to create word lattice, then compose word lattice with language model to obtain word lattice with combined acoustic/language weights.

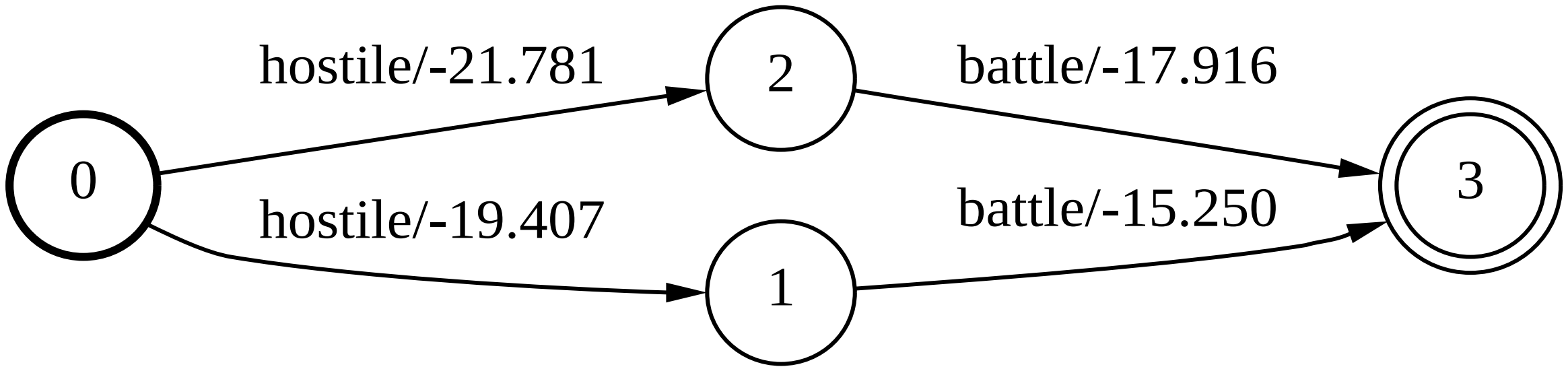

## Language Identification

- Language identification can be approached by simultaneously recognizing in $N$ languages and selecting the language with the best recognition score.
- In weakly-constrained task domains, the combined “lexicon” and “language model” for each language may need to be correspondingly weak (but general) – e.g., phone or syllable n-grams.
- In more strongly-constrained task domains, more lexical and grammatical/semantic constraints can be used as in conventional ASR systems, e.g., ranging from word-spotting to full trigram language models.
- Constructing such systems requires multilingual text normalization and pronunciation components for training and testing.